\documentclass[10pt,twocolumn,letterpaper]{article}

\usepackage{cvpr}
\usepackage{times}
\usepackage{epsfig}
\usepackage{graphicx}
\usepackage{amsmath}
\usepackage{amssymb}

\usepackage{multirow}
\usepackage{makecell}

\usepackage[breaklinks=true,bookmarks=false]{hyperref}

\cvprfinalcopy 

\def\cvprPaperID{7575} 

\ifcvprfinal\fi
\begin{document}

\title{Unsupervised Person Re-identification via Multi-label Classification}

\author{Dongkai Wang \ \quad Shiliang Zhang\\
Peking University\\
{\tt\footnotesize \{dongkai.wang, slzhang.jdl\}@pku.edu.cn}
}

\maketitle
\thispagestyle{empty}

\begin{abstract}
   The challenge of unsupervised person re-identification (ReID) lies in learning discriminative features without true labels. This paper formulates unsupervised person ReID as a multi-label classification task to progressively seek true labels. Our method starts by assigning each person image with a single-class label, then evolves to multi-label classification by leveraging the updated ReID model for label prediction. The label prediction comprises similarity computation and cycle consistency to ensure the quality of predicted labels. To boost the ReID model training efficiency in multi-label classification, we further propose the memory-based multi-label classification loss (MMCL). MMCL works with memory-based non-parametric classifier and integrates multi-label classification and single-label classification in a unified framework. Our label prediction and MMCL work iteratively and substantially boost the ReID performance. Experiments on several large-scale person ReID datasets demonstrate the superiority of our method in unsupervised person ReID. Our method also allows to use labeled person images in other domains. Under this transfer learning setting, our method also achieves state-of-the-art performance.
\end{abstract}

\section{Introduction}
\begin{figure}
\begin{center}
\includegraphics[width=1\linewidth]{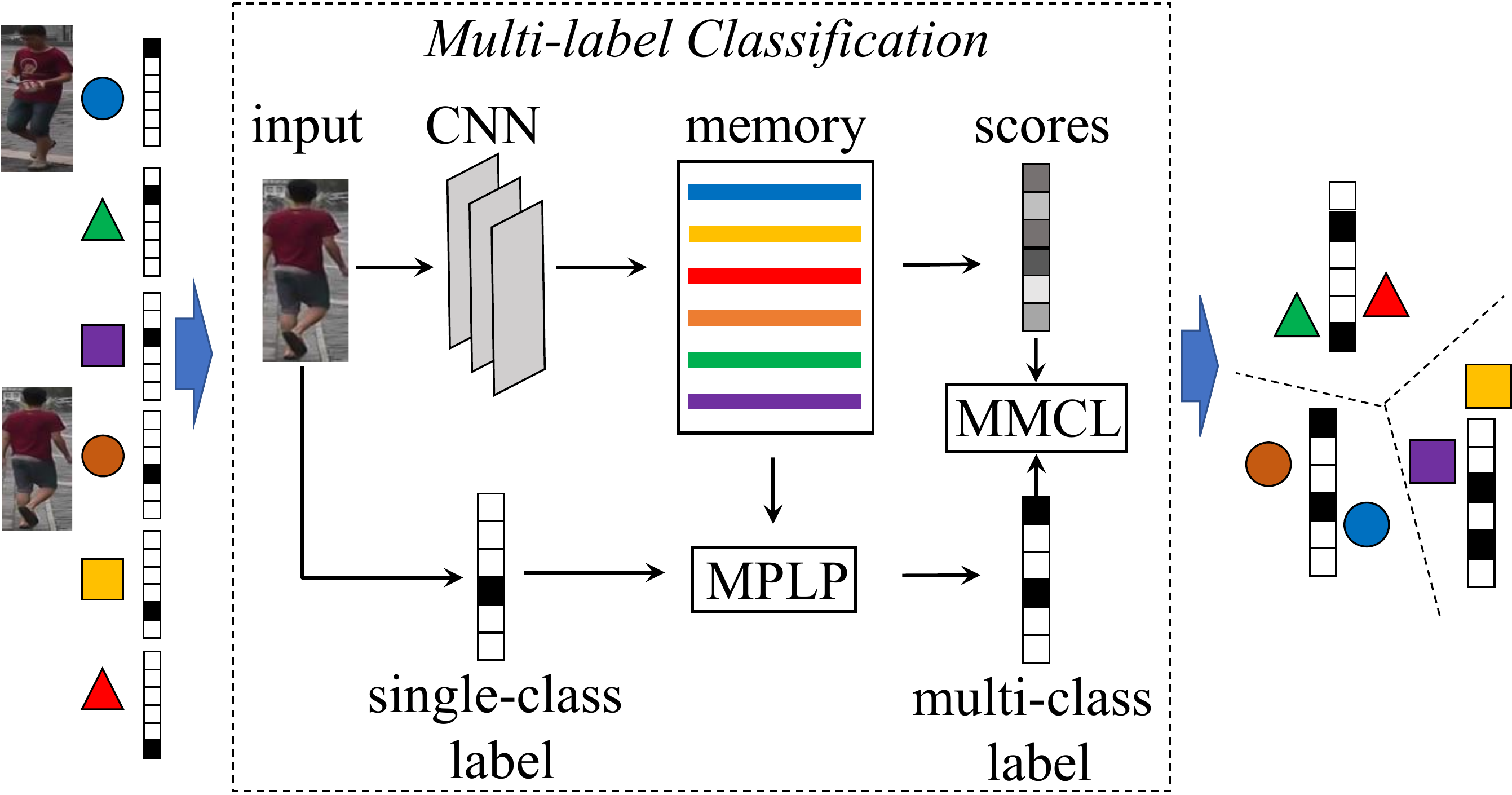}
\end{center}
\caption{Illustrations of multi-label classification for unsupervised person ReID. We target to assign each unlabeled person image with a multi-class label reflecting the person identity. This is achieved by iteratively running MPLP for prediction and MMCL for multi-label classification loss computation. This procedure guides CNN to produce discriminative features for ReID.}
\label{fig:idea}
\end{figure}

Recent years have witnessed the great success of person re-identification (ReID), which learns discriminative features from labeled person images with deep Convolutional Neural Network (CNN)~\cite{zheng2016person, krizhevsky2012imagenet, he2016deep,su2015multi,li2019pose,su2017pose,su2016deep,su2017attributes}. Because it is expensive to annotate person images across multiple cameras, recent research efforts start to focus on unsupervised person ReID. Unsupervised person ReID aims to learn discriminative features from unlabeled person images. Compared with supervised learning, unsupervised learning relieves the requirement for expensive data annotation, hence shows better potential to push person ReID towards real applications.


The challenge of unsupervised person ReID lies in learning discriminative features without true labels. To conquer this challenge, most of recent works~\cite{yu2017cross, zhong2019invariance, Lv_2018_CVPR, lin2018multi,wang2018transferable} define unsupervised person ReID as a transfer learning task, which leverages labeled data on other domains for model initialization or label transfer. Among them, some works assign each image with a single-class label~\cite{zhong2019invariance}. Some others leverage spatio-temporal cues or additional attribute annotations~\cite{Lv_2018_CVPR,lin2018multi,wang2018transferable}. Detailed review of existing methods will be presented in Sec.~\ref{sec:relatedwork}. Thanks to the above efforts, the performance of unsupervised person ReID has been significantly boosted. However, there is still a considerable gap between supervised and unsupervised person ReID. Meanwhile, the setting of transfer learning leads to limited flexibility. For example, as discussed in many works~\cite{long2015learning,Yan_2017_CVPR,wei2018person}, the performance of transfer learning is closely related to the domain gap, \emph{e.g.}, large domain gap degrades the performance. It is non-trivial to estimate the domain gap and select suitable source datasets for transfer learning in unsupervised person ReID.

This paper targets to boost unsupervised person ReID without leveraging any labeled data. As illustrated in Fig.~\ref{fig:idea}, we treat each unlabeled person image as a class and train the ReID model to assign each image with a multi-class label. In other words, the ReID model is trained to classify each image to multiple classes belonging to the same identity. Because each person usually has multiple images, multi-label classification effectively identifies images of the same identity and differentiates images from different identities. This in-turn facilitates the ReID model to optimize inter and intra class distances. Compared with previous methods~\cite{lin2019bottom,wu2018unsupervised}, which classify each image into a single class, the multi-label classification has potential to exhibit better efficiency and accuracy.

Our method iteratively predicts multi-class labels and updates the network with multi-label classification loss. As shown in Fig.~\ref{fig:idea}, to ensure the quality of predicted labels, we propose the Memory-based Positive Label Prediction (MPLP), which considers both visual similarity and cycle consistency for label prediction. Namely, two images are assigned with the same label if they a) share large similarity and b) share similar neighbors. To further ensure the accuracy of label prediction, MPLP utilizes image features stored in the memory bank, which is updated with augmented features after each training iteration to improve feature robustness.

Predicted labels allow for CNN training with a multi-label classification loss. Since each image is treated as a class, the huge number of classes makes it hard to train classifiers like Fully Connected (FC) layers. As shown in Fig.~\ref{fig:idea}, we adopt the feature of each image stored in the memory bank as a classifier. Specifically, a Memory-based Multi-label Classification Loss (MMCL) is introduced. MMCL accelerates the loss computation and addresses the vanishing gradient issue in traditional multi-label classification loss~\cite{zhang2013review,Durand_2019_CVPR} by abandoning the sigmoid function and enforcing the classification score to 1 or -1. MMCL also involves hard negative class mining to deal with the imbalance between positive and negative classes.

We test our approach on several large-scale person ReID datasets including Market-1501~\cite{zheng2015scalable}, DukeMTMC-reID~\cite{ristani2016performance} and MSMT17~\cite{wei2018person} without leveraging other labeled data. Comparison with recent works shows our method achieves competitive performance. For instance, we achieve rank-1 accuracy of 80.3\% on Market-1501, significantly outperforming the recent BUC~\cite{lin2019bottom} and DBC~\cite{ding12dispersion} by 14.1\% and 11.1\%, respectively. Our performance is also better than the 
HHL~\cite{zhong2018generalizing} and ECN~\cite{zhong2019invariance}, which use extra DukeMTMC-reID~\cite{ristani2016performance} for transfer learning. Our method is also compatible with transfer learning. Leveraging DukeMTMC-reID for training, we further achieve rank-1 accuracy of 84.4\% on Market-1501.

In summary, our method iteratively runs MPLP and MMCL to seek true labels for multi-label classification and CNN training. As shown in our experiments, this strategy, although does not leverage any labeled data, achieves promising performance. The maintained memory bank reinforces both label prediction and classification. Our work also shows that, unsupervised training has potential to achieve better flexibility and accuracy than existing transfer learning strategies.

\section{Related Work} \label{sec:relatedwork}
This section briefly reviews related works on unsupervised person ReID, unsupervised feature learning, and multi-label classification.

\emph{Unsupervised person ReID} works can be summarized into three categories. The first category utilizes hand-craft features~\cite{liao2015person,zheng2015scalable}. However, it is difficult to design robust and discriminative features by hand. The second category~\cite{fan2018unsupervised,Fu_2019_ICCV} adopts clustering to estimate pseudo labels to train the CNN. However, these methods require good pre-trained model. Apart from this, Lin \emph{et al.}~\cite{lin2019bottom} treat each image as a cluster, through training and merging clusters, this method achieves good performance.

The third category~\cite{lin2018multi,wang2018transferable,yu2019unsupervised,wei2018person,deng2018image,zhong2018generalizing,zhong2019invariance,
Chen_2019_ICCV,Wu_2019_ICCV,Li_2019_ICCV,Qi_2019_ICCV,Zhang_2019_ICCV} utilizes transfer learning to improve unsupervised person ReID. Some works~\cite{lin2018multi,wang2018transferable} use transfer learning and minimize the attribute-level discrepancy by utilizing extra attribute annotations. MAR~\cite{yu2019unsupervised} uses the source dataset as a reference to learn soft labels, which hence supervise the ReID model training. Generative Adversarial Network (GAN) is also utilized for transfer learning. PTGAN~\cite{wei2018person} and SPGAN~\cite{deng2018image} first generate transferred images from source datasets, then uses transferred image for training. HHL~\cite{zhong2018generalizing} generates images under different cameras and trains network using triplet loss. ECN~\cite{zhong2019invariance} utilizes transfer learning and minimizes the target invariance. Transfer learning requires a labeled source dataset for training. Our method differs with them that, it does not require any labeled data. It also achieves better performance than many transfer learning methods.

\emph{Unsupervised feature learning} aims to relieve the requirement on labeled data for feature learning. It can be applied in different tasks. Some works adopt unsupervised feature learning for neural network initialization. For example, RotNet~\cite{komodakis2018unsupervised} predicts the rotation of image to learn a good representation. Li \emph{et al.}~\cite{li2018unsupervised} use motion and view as supervision to learn an initialization for action recognition. Some other works utilize unsupervised feature learning to acquire features for image classification and retrieval. \cite{iscen2018mining} utilizes manifold learning to seek positive and negative samples to compute the triplet loss. Wu \emph{et al.}~\cite{wu2018unsupervised} regard each image as single class, and propose a non-parametric softmax classifier to train CNN. Our work shares certain similarity with \cite{wu2018unsupervised}, in that we also use non-parametric classifiers. However, we consider multi-label classification, which is important in identifying images of the same identity as well as differentiating different identities.

\emph{Multi-label classification} is designed for classification tasks with multi-class labels~\cite{zhang2013review,Durand_2019_CVPR,wang2018transferable,lin2018multi}. Durand \emph{et al.}~\cite{Durand_2019_CVPR} deal with multi-label learning based on partial labels and utilize GNN to predict missing labels. Wang \emph{et al.}~\cite{wang2018transferable,lin2018multi} use multi-label classification to learn attribute feature. This paper utilizes multi-label classification to predict multi-class labels and focuses on learning identity feature for person ReID. To the best of our knowledge, this is an early work utilizing multi-label classification for unsupervised person ReID.

\section{Methodology}
\subsection{Formulation} ~\label{sec:formulate}
Given an unlabeled person image dataset $\mathcal X=\{x_1,x_2,...,x_n\}$, our goal is to train a person ReID model on $\mathcal X$. For any query person image $q$, the person ReID model is expected to produce a feature vector to retrieve image $g$ containing the same person from a gallery set $G$. In other words, the ReID model should guarantee $q$ share more similar feature with $g$ than with other images in $G$. We could conceptually denote the goal of person ReID as,
\begin{equation}
g^* = \arg\min_{g\in G} \operatorname {dist}(f_g,f_q),
\end{equation}
where $f \in \mathbb{R}^d$ is a $d$-dimensional L2-normalized feature vector extracted by the person ReID model. $\operatorname {dist}(\cdot)$ is the distance metric, \emph{e.g.}, the L2 distance.

To make training on $\mathcal X$ possible, we start by treating each image as an individual class and assign $x_i$ with a label $y_i$. This pseudo label turns $\mathcal X$ into a labeled dataset, and allows for the ReID model training. $y_i$ is initialized to a two-valued vector, where only the value at index $i$ is set to $1$ and the others are set to $-1$, \emph{i.e.},
\begin{equation}
y_i[j]=\left\{
\begin{array}{rcl}
1 & & {j = i}\\
-1 & & {j \neq i}
\end{array} \right.
\end{equation}

Since each person may have multiple images in $\mathcal X$, the initial label vector is not valid in representing person identity cues. Label prediction is required to assign multi-class labels to each image, which can be used for ReID model training with a multi-label classification loss. Labels of $x_i$ can be predicted by referring its feature $f_i$ to features of other images, and find consistent feature groups. On the other hand, due to the huge number of image classes in $\mathcal X$, it is hard to train a multi-label classifier. One efficient solution is to use the $f_i$ as the classifier for the $i$-th class. This computes the classification score for any image $x_j$ as,
\begin{equation}
c_j[i]= f_i^\top \times f_j,
\end{equation}
where $c_j$ denotes the multi-label classification score for $x_j$.

It is easy to infer that, both label prediction and multi-label classification require features of images in $\mathcal X$. We hence introduce a $n \times d$ sized memory bank $\mathcal{M}$ to store those features, where $\mathcal{M}[i] =f_i$. With $\mathcal{M}$, we propose the Memory-based Positive Label Prediction
(MPLP) for label prediction and Memory-based Multi-label Classification Loss (MMCL) for ReID model training, respectively.

As shown in Fig.~\ref{fig:idea}, MPLP takes a single-class label as input and outputs the multi-label prediction $\bar y_i$ based on memory bank $\mathcal{M}$, \emph{i.e.},
\begin{equation}~\label{eq:mplp}
\bar y_i = \operatorname {MPLP}(y_i, \mathcal{M}),
\end{equation}
where $\operatorname {MPLP}(\cdot)$ denotes the MPLP module and $\bar y$ is the multi-class label.

MMCL computes the multi-label classification loss by taking the image feature $f$, label $\bar y$, and the memory bank $\mathcal{M}$ as inputs. The computed loss $\mathcal{L}_{MMCL}$ can be represented as,
\begin{equation}\label{eq:mmcl}
\mathcal{L}_{MMCL} = \sum_{i=1}^{n} \mathcal D(\mathcal{M}^\top \times f_i, \bar y_i),
\end{equation}
where $\mathcal{M}^\top \times f_i$ computes the classification score, and $\mathcal D(\cdot)$ computes the loss by comparing classification scores and predicted labels.

$\mathcal{M}$ is updated after each training iteration as,
\begin{equation}
\mathcal{M}[i]^t = \alpha \cdot f_i + (1-\alpha) \cdot \mathcal{M}[i]^{t-1},
\end{equation}
where the superscript $t$ denotes the $t$-th training epoch, $\alpha$ is the updating rate. $\mathcal{M}[i]^t$ is then L2-normalized by $\mathcal{M}[i]^t \leftarrow||\mathcal{M}[i]^t||_2$. It is easy to infer that, both MPLP and MMCL require robust features in $\mathcal{M}$ to seek reliable labels and classification scores, respectively. We use many data argumentation techniques to reinforce $\mathcal{M}$. In other words, each $\mathcal{M}[i]$ combines features of different augmented samples form $x_i$, hence it presents better robustness. More details are given in Sec.~\ref{sec:implement}.

MPLP considers both similarity and cycle consistency to predict $\bar y_i$, making it more accurate than the classification score. This makes the loss computed with Eq.~\eqref{eq:mmcl} valid in boosting the ReID model, which in-turn produces positive feedbacks to $\mathcal{M}[i]$ and label prediction. This loop makes it possible to train discriminative ReID models on unlabeled dataset. Implementations to MPLP and MMCL can be found in the following parts.

\subsection{Memory-based Positive Label Prediction}\label{section:mplp}

As shown in Eq.~\eqref{eq:mplp}, given an initial two-valued label $y_i$ of image $x_i$, MPLP aims to find other classes that $x_i$ may belong to. For $x_i$, MPLP first computes a rank list ${R}_{i}$ according to the similarity between $x_i$ and other features, \emph{i.e.},
\begin{equation}\label{eq:labelrank}
{R}_{i} = \mathop{\arg \operatorname {sort}}_j (s_{i,j}), j \in [1,n],
\end{equation}
\begin{equation}\label{eq:mem_sim}
s_{i,j} = \mathcal{M}[i]^\top \times \mathcal{M}[j],
\end{equation}
where $s_{i,j}$ denotes the similarity score of $x_i$ and $x_j$.

${R}_{i}$ finds candidates for reliable labels for $x_i$, \emph{e.g.}, labels at the top of rank list. However, variances of illumination, viewpoint, backgrounds, \emph{etc}., would degrade the robustness of the rank list. \emph{E.g.}, noisy labels may appear at the top of rank list. To ensure the quality of predicted labels, MPLP refers to the similarity score and cycle consistency for label prediction.

\begin{figure}
\begin{center}
\includegraphics[scale=0.19]{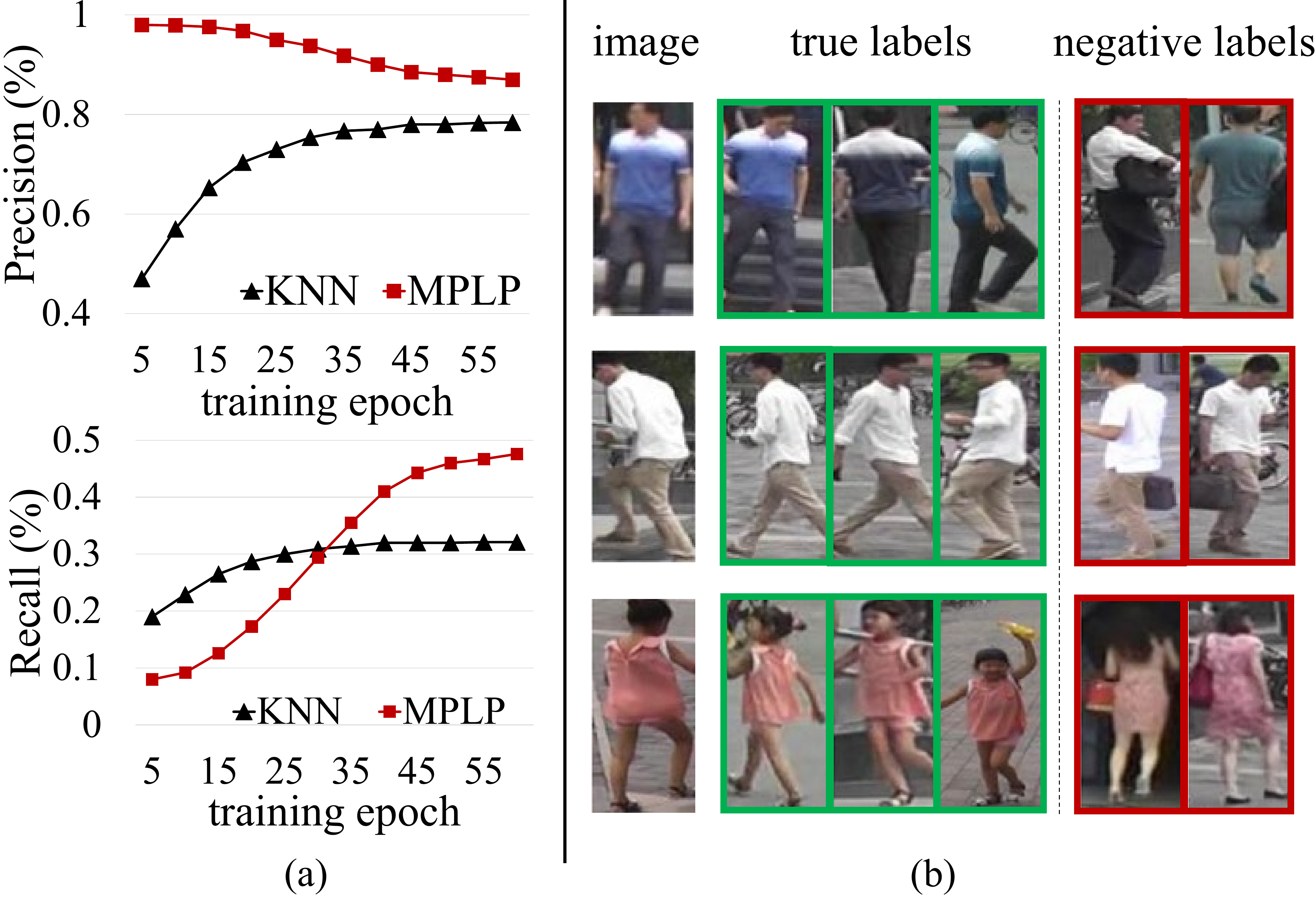}
\end{center}
 \caption{Illustration of the label prediction by MPLP. (a) reports the precision and recall of MPLP in finding true positive labels, where MPLP consistently outperforms KNN at different training stages. (b) shows positive labels and negative labels selected by MPLP, where MPLP effectively rejects hard negative labels.}
\vspace{-1mm}
\label{fig:visualization}
\end{figure}

\textbf{Label filtering by similarity score:} We first select positive label candidates for $x_i$ based on its rank list. Inspired by~\cite{zhang2013review}, that uses a threshold to select relevant labels for query, we select candidate labels with a predefined similarity threshold. Given a similarity score threshold $t$, $k_i$ label candidates can be generated by removing labels with similarity smaller than $t$, \emph{i.e.},
\begin{equation}
\begin{aligned}\label{eq:t_select}
{P}_{i} = {R}_{i}[1:k_i],
 \end{aligned}
\end{equation}
where ${R}_{i}[k_i]$ is the last label with similarity score higher than $t$, ${P}_{i}$ is the collection of label candidates for $x_i$. $t$ largely decides the quantity of label candidates. It will be tested in Sec.~\ref{section:pa}. Eq.~\eqref{eq:t_select} adaptively finds different numbers of candidates labels for different images, which is better than selecting fixed number of labels, \emph{i.e.}, the KNN in Fig.~\ref{fig:visualization}. We proceed to introduce cycle consistency to find positive labels from ${P}_{i}$.


\textbf{Label filtering by cycle consistency:} Inspired by k-reciprocal nearest neighbor\cite{jegou2007contextual,zhong2017re}, we assume that, if two images belong to the same class, their neighbor image sets should also be similar. In other words, two images should be mutual neighbor for each other if they can be assigned with similar labels. With this intuition, we propose a cycle consistency scheme to filter hard negative labels in $P_i$.

MPLP traverses labels in $P_i$ from head to tail. For a label $j$ in ${P}_{i}$, MPLP computes its top-$k_i$ nearest labels with Eq.~\eqref{eq:labelrank}. If label $i$ is also one of the top-$k_i$ nearest labels of $j$, $j$ is considered as a positive label for $x_i$. Otherwise, it is treated as a hard negative label. The traverse is stopped when the first hard negative label is found. This leads to a positive label set $P^*_i$ as well as a hard negative label for image $x_i$. We denote the positive label set as,
\begin{equation}
{P}^*_{i} = {P}_{i}[1:l],
\end{equation}
where $l$ satisfies $i \in {R}_{{P_i[l]}}[1:k_i] ~\&~ i \notin {R}_{{P_i[l+1]}}[1:k_i]$. 
As ${P}^*_{i}$ contains $l$ labels, $x_i$ would be assigned with a multi-class label $\bar y_i$ with $l$ positive classes,
\begin{equation}
\bar y_i[j]=\left\{
\begin{array}{rcl}
1 & & {j\in {P}^*_{i}}\\
-1 & & {j \notin {P}^*_{i}}
\end{array} \right.
\end{equation}

As Fig.~\ref{fig:visualization} shows that, MPLP predicts accurate positive labels. Experimental evaluations will be presented in Sec.~\ref{section:ablation}.

\subsection{Memory-based Multi-label Classification Loss}\label{section:mmcl}
\textbf{Traditional multi-label classification loss:} The predicted multi-class labels are used for training the ReID model with a multi-label classification loss. In traditional multi-label classification methods, sigmoid and logistic regression loss is a common option~\cite{zhang2013review,Durand_2019_CVPR,wang2018transferable,lin2018multi}. For a task with $n$ classes, it adopts $n$ independent binary classifiers for classification. The loss of classifying image $x_i$ to class $j$ can be computed as,
\begin{equation}~\label{eq:mcl}
\ell(j|x_i) = \log (1 + \exp (-\bar y_i[j]\times \mathcal{M}[j]^\top \times f_i)),
\end{equation}
where $\mathcal{M}[j]^\top \times f_i$ computes the classification score of image $x_i$ for the class $j$. $\bar y_i[j]$ is the label of image $x_i$ for class $j$. With the loss at a single class, we can obtain the Multi-Label Classification (MCL) loss, \emph{i.e.}, $\mathcal{L}_{MCL}$,
\begin{equation}
\mathcal{L}_{MCL} = \sum_{i=1}^{n}\sum_{j=1}^{n}\ell(j|x_i),
\end{equation}
where $n$ is the number of images in the dataset $\mathcal X$, which equals to the class number in our setting.

\begin{figure}
\begin{center}
\includegraphics[scale=0.25]{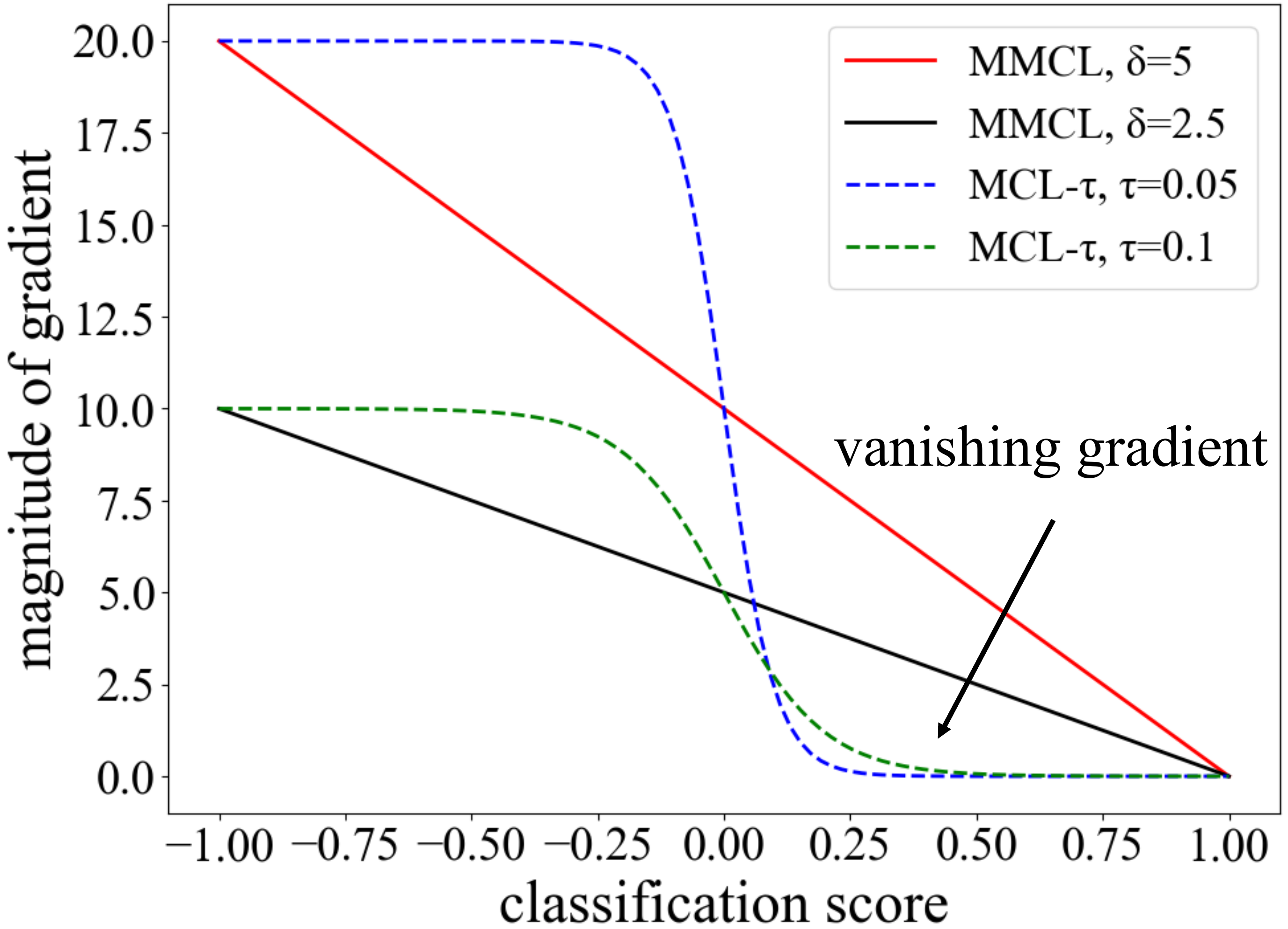}
\end{center}
   \caption{Gradient Analysis for MCL-$\tau$ and MMCL. It is clear that, MMCL does not suffer from the vanishing gradient issue.}
   \vspace{-2mm}
\label{fig:gradientanalysis}
\end{figure}

Because the $\mathcal{M}[j]^\top$ and $f_i$ are L2 normalized, the classification score is restricted between $[-1,1]$. This limits the range of sigmoid function in Eq.~\eqref{eq:mcl}, making the loss non-zero even for correct classifications. This issue can be addressed by introducing a scalar $\tau$ on the classification score. This updates Eq.~\eqref{eq:mcl} as,
\begin{equation}~\label{eq:mcl-t}
\ell_\tau(j|x_i) = \log (1 + \exp (-\bar y_i[j]\times \mathcal{M}[j]^\top \times f_i/\tau)).
\end{equation}
We denote the corresponding MCL loss as $\mathcal{L}_{MCL-\tau}$. The gradient of $\mathcal{L}_{MCL-\tau}$ can be computed as,
\begin{equation}~\label{eq:grad_mcl}
\begin{aligned}
\frac{\partial{\mathcal{L}_{MCL-\tau}}}{\partial{f_i}} = -\frac{\exp(-\bar y_i[j] \mathcal{M}[j]^\top f_i /\tau)}{1+\exp(-\bar y_i[j]\mathcal{M}[j]^\top f_i/\tau)}\frac{\bar y_i[j]\mathcal{M}[j]}{\tau}.
 \end{aligned}
\end{equation}

With Eq.~\eqref{eq:grad_mcl}, we illustrate the gradient of $\mathcal{L}_{MCL-\tau}$ with different values of $\tau$ when $\bar y_i[j]=1$ in Fig.~\ref{fig:gradientanalysis}. It is clear that, the updated MCL loss still suffers from substantial vanishing gradient issue as the classification score larger than 0.25 or smaller than -0.25.

Another issue with MCL loss is that, our task involves a large number of classes, making the positive and negative classes unbalanced. Treating those negative classes equally in Eq.~\eqref{eq:mcl-t} may cause a model collapse. We hence proceed to propose MMCL to address those issues.

\textbf{Memory-based Multi-label Classification Loss:} MMCL is proposed to address two issues in traditional MCL. For the first issue, since the score is bounded by $[-1,1]$, we can abandon the sigmoid function and directly compute the loss by regressing the classification score to 1 and -1. This simplifies the loss computation and improves the training efficiency. The loss of classifying image $x_i$ to class $j$ can be updated as,
\begin{equation}
\ell^*(j|x_i) = ||\mathcal{M}[j]^\top \times f_i -\bar y_i[j]||^2,
\end{equation}
where $f_i$ is the feature of image $x_i$.

The second problem is the imbalance between positive and negative classes. MMCL introduces hard negative class mining to solve it. This is inspired by the sample mining in deep metric learning~\cite{wu2017sampling}, where hard negative samples are more informative for training. Similarly in our multi-label classification, the training should focus more on hard negative classes than easy negative classes.

For $x_i$, its negative classes can be denoted as $R_{i} \backslash P^*_{i}$. We rank them by their classification scores and select the top $r\%$ classes as the hard negative classes. The collection of hard negative classes for $x_i$ can be denoted as ${N}_{i}, |{N}_{i}|=(n-|P^*_{i}|)\cdot r\%$.

The MMCL is computed on positive classes and sampled hard negative classes as follows, 
\begin{equation}\label{mmcl}
\begin{aligned}
\mathcal{L}_{MMCL} = \sum_{i=1}^{n} \frac{\delta}{|{P}^*_{i}|}\sum_{p \in \mathcal{P}^*_{i}}\ell^*(p|x_i) + \\
 \frac{1}{|{N}_{i}|}\sum_{s \in {N}_{i}}\ell^*(s|x_i)
 \end{aligned}
\end{equation}
where $\delta$ is a coefficient measuring the importance of positive class loss and negative class loss, which will be tested in experiments.

We also illustrate the gradients of $\mathcal{L}_{MMCL}$ similarly when $\bar y_i[j]=1$, in Fig.~\ref{fig:gradientanalysis}, where the gradient of $\mathcal{L}_{MMCL}$ can be computed as,
\begin{equation}
\begin{aligned}
\partial{\mathcal{L}_{MMCL}}/{\partial{f_i}}= 2\delta (\mathcal{M}[j]^\top \times f_i -\bar y_i[j])\mathcal{M}[j].
 \end{aligned}
\end{equation}

\textbf{Discussions:} Comparison between MCL and MMCL in Fig.~\ref{fig:gradientanalysis} clearly shows that, the vanishing gradient issue is effectively addressed by MMCL. Because of vanishing gradient, $\mathcal{L}_{MCL-\tau}$ won't enforce the classifier to classify positive labels with large scores. This is harmful for decreasing the intra-class variance. Therefore, MMCL is more effective than MCL in optimizing the ReID model. Fig.~\ref{fig:gradientanalysis} also shows that, $\delta$ controls the magnitude of the gradient of MMCL. As discussed in \cite{NIPS2018_8094}, mean square loss is inferior to log-based loss (\emph{e.g.} cross entropy) when classification score is near the decision boundary. $\delta$ effectively solves this issue by scaling the gradient magnitude of MMCL.

By adopting the hard negative class mining strategy, MMCL not only works for multi-label classification, it also could be applied in single-label classification, where the unbalanced class issue still exists. Compared with cross entropy loss and MCL, MMCL abandons activation functions like softmax and sigmoid, leading to more efficient computation. As discussed in many works~\cite{morin2005hierarchical, gutmann2010noise}, a huge number of classes degrades the speed of softmax computation. Existing solutions include hierarchical softmax~\cite{morin2005hierarchical} and noise-contrastive estimation~\cite{gutmann2010noise}. As MMCL does not involve softmax computation, it does not suffer from such issues.

\subsection{Transfer Learning with Labeled Dataset}\label{section:transfer}
Our method is also compatible with transfer learning setting. Given a dataset containing labeled person images, we can adopt the commonly used cross entropy loss and triplet loss on labeled data to train the model. The training loss can be denoted as $\mathcal{L}_{labeled}$. The overall training loss for transfer learning can be represented as the sum of MMCL and loss on labeled dataset, \emph{i.e.},
\begin{equation}
\mathcal {L}_{transfer} = \mathcal{L}_{labeled} + \mathcal{L}_{MMCL}.
\end{equation}
The performance of our methods on transfer learning will be tested in the next section.

\section{Experiments}
\vspace{-1mm}
\subsection{Dataset and Evaluation Metrics}
\vspace{-1mm}
\emph{Market-1501}~\cite{zheng2015scalable} contains 32,668 labeled person images of 1,501 identities collected from 6 non-overlapping camera views. \emph{DukeMTMC-reID}~\cite{ristani2016performance} has 8 cameras and 36,411 labeled images of 1,404 identities. \emph{MSMT17}~\cite{wei2018person} is a newly released person ReID dataset. It is composed of 126,411 person images from 4,101 identities collected by 15 cameras. The dataset suffers from substantial variations of scene and lighting, and is more challenging than the other two datasets. All three datasets are collected under similar scenario, \emph{i.e.}, campus, which makes the transfer learning possible. We follow the standard settings~\cite{zheng2015scalable,ristani2016performance,wei2018person} on them to conduct experiments. Performance is evaluated by the Cumulative Matching Characteristic (CMC) and mean Average Precision (mAP).

\begin{figure}
\begin{center}
\includegraphics[scale=0.24]{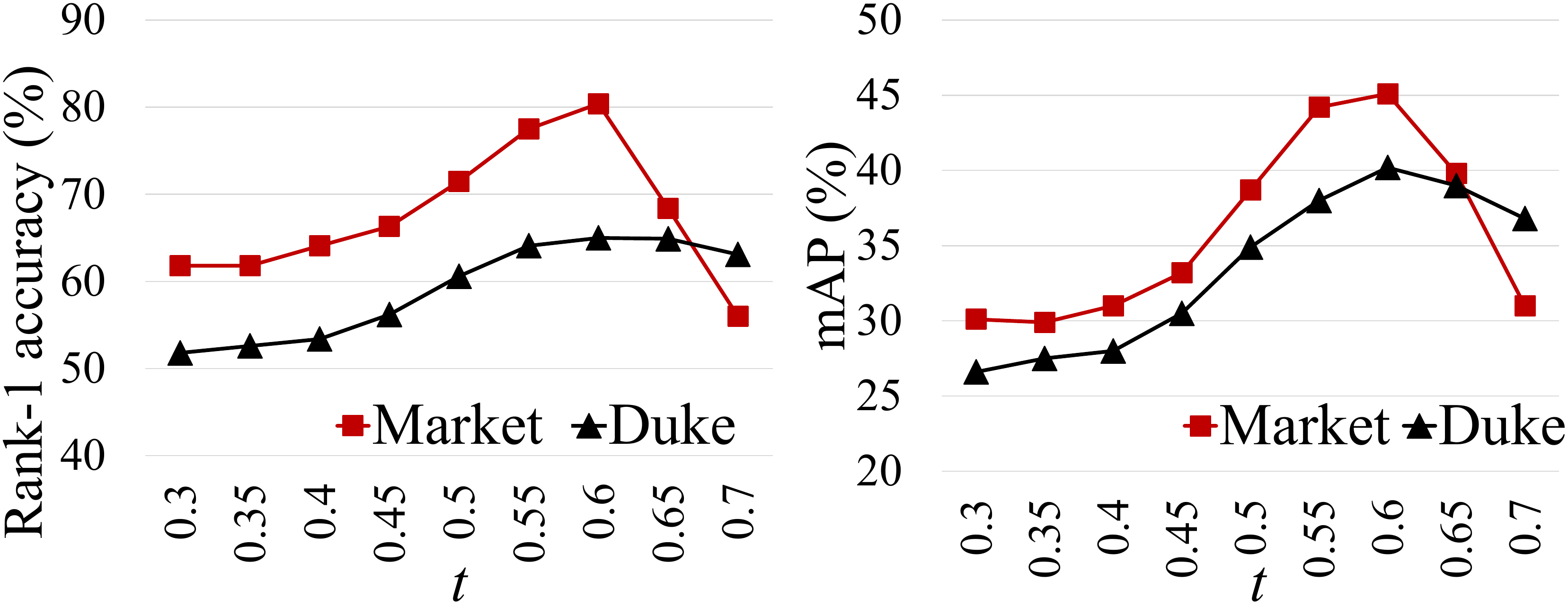}
\end{center}
\vspace{-2mm}
\caption{Evaluation of similarity score $t$ in MPLP.}
\vspace{-2mm}
\label{figure:pa-sst}
\end{figure}

\vspace{-1mm}
\subsection{Implementation Details}\label{sec:implement}
\vspace{-1mm}
All experiments are implemented on PyTorch. We use ResNet-50~\cite{he2016deep} as backbone to extract the feature and initialize it with parameters pre-trained on ImageNet~\cite{5206848}. After pooling-5 layer, we remove subsequent layers and add a batch normalization layer~\cite{ioffe2015batch}, which produces a 2048-dim feature. During testing, we also extract the pooling-5 feature to calculate the distance. For multi-label classification, we allocate a memory bank to store L2 normalized image features. The memory bank is initialized to all zeros, and we start using MPLP for label prediction when the memory is fully updated 5 times (after 5 epochs). As mentioned in section \ref{section:mplp}, we leverage CamStyle~\cite{zhong2018camera} as a data augmentation strategy for unlabeled images. Strategies like random crop, random rotation, color jitter, and random erasing are also introduced to improve the feature robustness.

The input image is resized to 256*128. We use SGD to optimize the model, the learning rate for ResNet-50 base layers are 0.01, and others are 0.1. The memory updating rate $\alpha$ starts from $0$ and grows linearly to $0.5$. We train the model for 60 epochs, and the learning rate is divided by 10 after every 40 epochs. The batch size for model training is 128. We fix the similarity threshold $t$ in MPLP as 0.6. In MMCL, the weight $\delta$ is fixed to 5 and we select the 1\% top-ranked negative classes to compute the loss through the parameter analysis in Sec.~\ref{section:pa}.

For transfer learning with labeled dataset, we apply the same batch size on the labeled dataset. A fully connected layer is added after batch normalization layer for classification. We optimize the $\mathcal {L}_{transfer}$ in section \ref{section:transfer} following the same baseline training strategy as described in \cite{Fu_2019_ICCV}.

\vspace{-1mm}
\subsection{Parameter Analysis}\label{section:pa}
\vspace{-1mm}
This section aims to investigate some important hyper-parameters in our method, including the similarity score threshold $t$ in MPLP, coefficient $\delta$, and hard negative mining ratio $r$\% in MMCL. Each experiment varies the value of one hyper-parameter while keeping others fixed. All experiments are conducted with unsupervised ReID setting on both Market-1501 and DukeMTMC-reID.

\begin{table}
\footnotesize
\begin{center}
\begin{tabular}{p{1cm}<{\centering}|cc|cc}
\hline
\multirow{2}{*}{$\delta$} & \multicolumn{2}{c|}{Market-1501} & \multicolumn{2}{c}{DukeMTMC-reID}\\
\cline{2-5}
 & Rank-1 & mAP & Rank-1 & mAP  \\
\hline\hline
1 & 59.3 & 19.4 & 52.6 & 22.8 \\
2 & 71.3 & 31.1 & 58.8 & 31.0 \\
3 & 76.6 & 40.0 & 62.6 & 35.6 \\
4 & 79.9 & 44.9 & 64.9 & 39.1 \\
5 & \textbf{80.3} & \textbf{45.5} & \textbf{65.2} & \textbf{40.2}\\
6 & 78.1 & 45.0 & 65.0 & 39.8\\
7 & 73.2 & 41.3 & 64.1 & 39.4\\
\hline
\end{tabular}
\end{center}
\vspace{-2mm}
\caption{Evaluation of parameter $\delta$ in Eq.~\eqref{mmcl}.}
\vspace{-2mm}
\label{table:pa-delta}
\end{table}

\textbf{Similarity threshold $t$:} Fig.\ref{figure:pa-sst} investigates the effect of similarity threshold $t$ in MPLP. We vary $t$ from $0.3$ to $0.7$ and test the model performance. A low similarity score $t$ will harm the model performance. For example, when $t$ is in range $[0.3,0.5]$, a substantial performance drop can be observed compared with larger $t$. This is because that, low similarity threshold introduces many negative labels. More accurate labels can be selected as $t$ becomes larger. However, too large $t$ decreases the number of selected labels. The best $t$ is 0.6 for both Market-1501 and DukeMTMC-reID. We hence set $t=0.6$.

\textbf{Coefficient $\delta$:} Table~\ref{table:pa-delta} reports the analysis on coefficient $\delta$ of MMCL. As discussed in Sec.~\ref{section:mmcl}, $\delta$ plays a role to scale the gradient of MMCL. $\delta=1$ means that we do not scale the gradient. In this case, the MMCL cannot produce large gradients to pull positive samples together, leading to bad performance. For example, the rank-1 accuracy is dropped to 59.3\% on Market-1501 and 52.6\% on DukeMTMC-reID. As $\delta$ becomes larger, MMCL effectively improves the similarity of positive samples, leading to better performance. However, too large $\delta$ may make the training unstable. According to Table~\ref{table:pa-delta}, we set $\delta=5$.

\begin{figure}
\begin{center}
\includegraphics[scale=0.25]{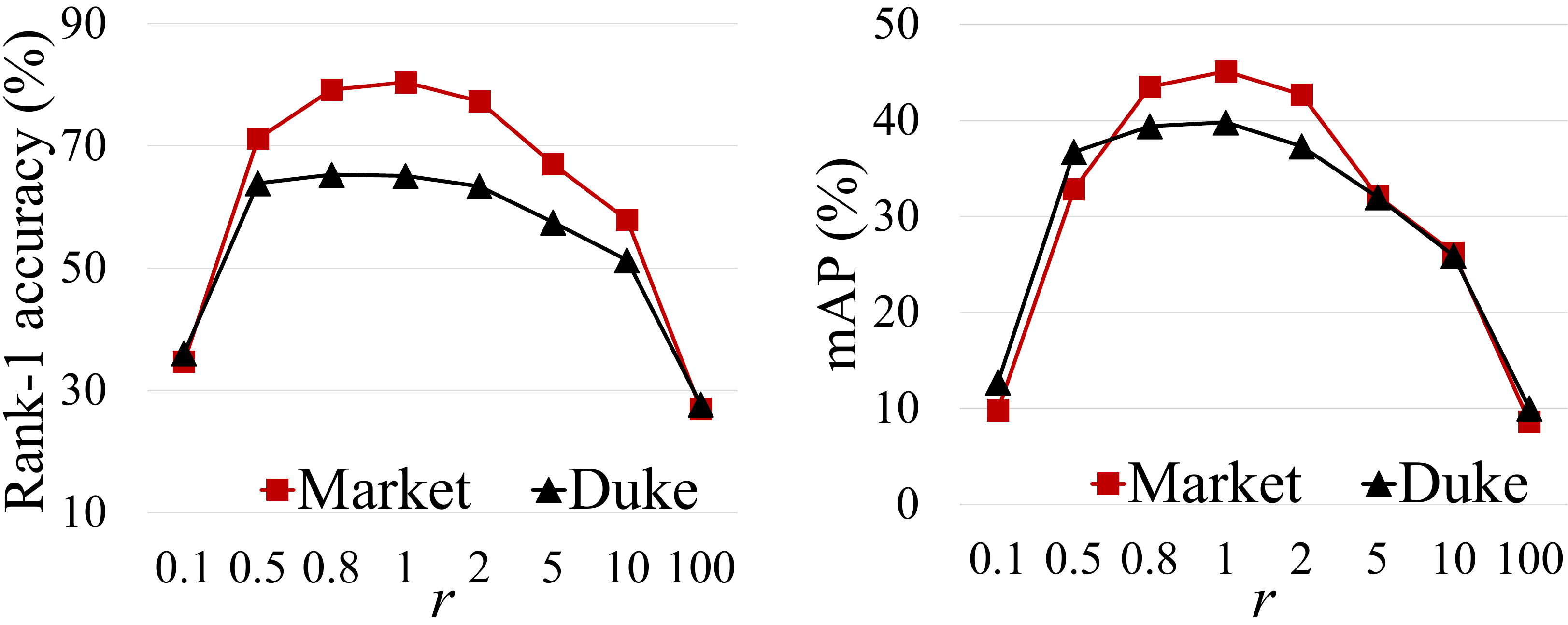}
\end{center}
\vspace{-2mm}
\caption{Evaluation of hard negative mining ratio $r$ in MMCL.}
\label{figure:pa-hnmr}
\end{figure}

\textbf{Hard negative mining ratio $r$\%:} Fig.~\ref{figure:pa-hnmr} shows effects of hard negative mining ratio $r$\% in network training. $r=100$ means using all negative classes for loss computation. It is clear that, $r=100$ is harmful for the performance. This implies that, not all of the negative classes are helpful for unsupervised ReID training. As $r$ becomes smaller, hard negative mining would be activated and it boosts the performance. Too small $r$ selects too few negative classes, hence is also harmful for the performance. According to Fig.~\ref{figure:pa-hnmr}, $r=1$ is used in our experiments.

\begin{table}
\footnotesize
\begin{center}
\begin{tabular}{l||c|c||c|c}
\hline
\multirow{2}{*}{Method} & \multicolumn{2}{c||}{Market-1501} & \multicolumn{2}{c}{DukeMTMC-reID}\\
\cline{2-5}
 & Rank-1 & mAP & Rank-1 & mAP  \\
\hline\hline
Supervised & 87.1 & 68.3 & 75.9 & 57.9 \\
\hline
ImageNet Pre-trained & 7.8 & 2.1 & 5.1 & 1.4 \\
MMCL + Single Label  & 49.0 & 17.8 & 42.0 & 16.6\\
MMCL + MPLP$\dagger$ & {66.6} & {35.3} & {58.0} & {36.3} \\
MMCL + MPLP & \textbf{80.3} & \textbf{45.5} & \textbf{65.2} & \textbf{40.2} \\
\hline
\end{tabular}
\end{center}
\vspace{-2mm}
\caption{Test of validity of MMCL and MPLP. $\dagger$ denotes not using CamStyle~\cite{zhong2018camera} for data augmentation.}
\vspace{-2mm}
\label{table:ablation}
\end{table}

\vspace{-1mm}
\subsection{Ablation Study}\label{section:ablation}
\vspace{-1mm}
This part evaluates the effectiveness of MPLP and MMCL by making comparison with supervised learning, MMCL+single-class label, and MMCL+MPLP. Experimental results are reported in Table \ref{table:ablation}, where the performance of ImageNet pre-trained model is also reported as the baseline. As shown in the table, the supervised learning achieves high accuracy, \emph{e.g.}, 87.1\% in rank-1 accuracy and 68.3\% in mAP on Market-1501. ImageNet pre-trained model performs badly on both Market-1501 and DukeMTMC-reID. MMCL with single-class pseudo labels boosts the baseline performance, indicating the validity of leveraging unlabeled dataset in training. Table \ref{table:ablation} also shows that, combining MMCL with MPLP significantly boosts the performance, \emph{e.g.}, from baseline 7.8\% to 80.3\% in rank-1 accuracy on Market-1501. Table \ref{table:ablation} also shows that, CamStyle~\cite{zhong2018camera} boosts the performance, indicating the importance of data augmentation as discussed in Sec.~\ref{sec:formulate}.

\textbf{Effectiveness of MPLP:} To verify that MPLP is a reasonably good solution for label prediction, we compare MPLP against several other label prediction methods, \emph{e.g.}, the KNN search and selection by Similarity Score (SS). Table \ref{table:labelandloss} (a) summarizes the results. From Table \ref{table:labelandloss} (a), we can observe that KNN (K=8, which achieves the best peformance) achieves 73.3\% rank-1 accuracy and 35.4\% mAP on Market-1501. 
Selecting positive labels by Similarity Score (SS) gains improvements over KNN. This indicates that, adaptively select positive labels with similarity threshold is more reasonable than fixing the positive label number for different images. Table \ref{table:labelandloss} (a) also shows that, MPLP achieves the best performance. This indicates that combining cycle consistency and similarity score effectively ensure the quality of predicted labels. Visualization of predicted labels by MPLP can be found in Fig.~\ref{fig:visualization}.

\begin{table}
\footnotesize
\begin{center}
\begin{tabular}{l||c|c||c|c}
\hline
\multirow{2}{*}{Method} & \multicolumn{2}{c||}{Market-1501} & \multicolumn{2}{c}{DukeMTMC-reID}\\
\cline{2-5}
 & Rank-1 & mAP & Rank-1 & mAP  \\
\hline\hline
MMCL + KNN &73.3 & 35.4 & 59.5 & 33.5 \\
MMCL + SS & 77.6 & 43.0 & 62.7 & 38.5 \\
MMCL + MPLP & \textbf{80.3} & \textbf{45.5} & \textbf{65.2} & \textbf{40.2} \\
\hline
\multicolumn{5}{c}{(a)}\\
\hline
CE + Single Label & 31.4 & 9.9 & 31.5 & 11.8 \\
MMCL + Single Label & 49.0 & 17.8 & 42.0 & 16.6\\
CE + Ground Truth & 85.5 & 65.5 & 70.4 & 50.3 \\
MMCL + Ground Truth & 86.5 & 67.2 & 74.5 & 56.9\\
CE + MPLP & 65.0& 34.2 & 60.2 & 35.1\\
MMCL + MPLP & 80.3 & 45.5 & 65.2 & 40.2 \\
\hline
\multicolumn{5}{c}{(b)}\\
\end{tabular}
\end{center}
\vspace{-2mm}
\caption{Ablation study on different label prediction algorithms and loss functions.}
\vspace{-2mm}
\label{table:labelandloss}
\end{table}

\textbf{Effectiveness of MMCL:} To test the validity of MMCL, this part proceeds to compare it against Cross Entropy (CE) loss with different training settings. Experimental results are summarized in Table \ref{table:labelandloss} (b). We first test MPLP and CE using single-class labels for model learning. MMCL gets 49.0\% rank-1 accuracy and 17.8\% mAP on Market-1501, substantially better than CE. We further test MMCL and CE using ground truth labels for learning. Note that, we modify CE according to \cite{zhong2019invariance} to make it applicable in multi-class learning. With ground truth labels, MMCL still performs better than CE, especially on DukeMTMC-reID. We finally test MMCL and CE with labels predicted by MPLP. MMCL still outperforms CE by large margins. Note that, MMCL uses non-parameters classifiers for training. It still achieves comparable performance with the supervised learning in Table \ref{table:ablation}. Table \ref{table:labelandloss} (b) thus demonstrates the effectiveness of MMCL and the training paradigm of our method.

\begin{table*}
\footnotesize
\begin{center}
\setlength{\tabcolsep}{2.5mm}{
\begin{tabular}{l||c||c|c|c|c|c||c|c|c|c|c}
\hline
\multirow{2}{*}{Method} & \multirow{2}{*}{Reference} & \multicolumn{5}{c||}{Market-1501} & \multicolumn{5}{c}{DukeMTMC-reID} \\
\cline{3-12}
&  & Source & Rank-1 & Rank-5 & Rank-10 & mAP & Source & Rank-1 & Rank-5 & Rank-10 & mAP \\
\hline\hline
LOMO\cite{liao2015person} & CVPR15& None &  27.2& 41.6& 49.1 &8.0& None & 12.3 &21.3& 26.6& 4.8 \\
BOW\cite{zheng2015scalable} & ICCV15&None & 35.8& 52.4& 60.3 &14.8 & None &17.1& 28.8& 34.9 &8.3 \\
BUC\cite{lin2019bottom} & AAAI19 & None & 66.2 & 79.6 & 84.5 & 38.3 & None& 47.4 & 62.6 & 68.4 & 27.5 \\
DBC\cite{ding12dispersion} & BMVC19 & None & 69.2 & 83.0 & 87.8 & 41.3 & None& 51.5 & 64.6 & 70.1 & 30.0 \\
\hline
Ours & This paper & None& \textbf{80.3} & \textbf{89.4} & \textbf{92.3} & \textbf{45.5} & None & \textbf{65.2} & \textbf{75.9} & \textbf{80.0} & \textbf{40.2}  \\
\hline
\hline
PUL\cite{fan2018unsupervised} & TOMM18 & Duke & 45.5 &60.7 &66.7& 20.5& Market & 30.0 &43.4 &48.5 &16.4 \\
PTGAN\cite{wei2018person} &CVPR18& Duke & 38.6& -& 66.1& -& Market & 27.4& - &50.7 & - \\
SPGAN\cite{deng2018image} & CVPR18& Duke & 51.5& 70.1 &76.8& 22.8& Market &41.1& 56.6& 63.0& 22.3 \\
CAMEL\cite{yu2017cross} & ICCV17& Multi & 54.5& -& -& 26.3&- & -& -& - & -\\
MMFA\cite{lin2018multi} & BMVC19 & Duke &56.7& 75.0& 81.8& 27.4& Market & 45.3& 59.8 &66.3& 24.7 \\
TJ-AIDL\cite{wang2018transferable} & CVPR18& Duke &58.2 &74.8 &81.1 &26.5& Market & 44.3& 59.6 &65.0& 23.0 \\
HHL\cite{zhong2018generalizing} &ECCV18 & Duke & 62.2 & 78.8 & 84.0 & 31.4  &Market &46.9 & 61.0 & 66.7 & 27.2 \\
ECN\cite{zhong2019invariance} & CVPR19 & Duke & 75.1 & 87.6 & 91.6 & 43.0 &Market & 63.3 & 75.8 & 80.4 & 40.4 \\
MAR\cite{yu2019unsupervised}  &CVPR19 & MSMT & 67.7  &81.9  &-  &40.0 & MSMT & 67.1 & 79.8  & -  &48.0\\
PAUL\cite{Yang_2019_CVPR} & CVPR19 & MSMT & 68.5 & 82.4 & 87.4 & 40.1 & MSMT & 72.0 & 82.7 & \textbf{86.0} & 53.2 \\
SSG\cite{Fu_2019_ICCV} & ICCV19 & Duke & \underline{80.0} & \underline{90.0} & \underline{92.4} & \underline{58.3} & Market & \textbf{73.0} & 80.6 & 83.2 & \underline{53.4} \\
CR-GAN\cite{Chen_2019_ICCV} & ICCV19 & Duke & 77.7 & 89.7 & 92.7 & 54.0 & Market & 68.9 & 80.2 & 84.7 & 48.6 \\
CASCL\cite{Wu_2019_ICCV} & ICCV19 & MSMT & 65.4 & 80.6 & 86.2 & 35.5 & MSMT & 59.3 & 73.2 & 77.5 & 37.8\\
PDA-Net\cite{ Li_2019_ICCV} & ICCV19 & Duke & 75.2 & 86.3 & 90.2 & 47.6 & Market & 63.2 & 77.0 & 82.5 & 45.1\\
UCDA\cite{Qi_2019_ICCV} & ICCV19 & Duke & 64.3 & - & - & 34.5 & Market & 55.4 & - & - & 36.7\\
PAST\cite{Zhang_2019_ICCV} & ICCV19 & Duke & 78.38 & - & - & 54.62 & Market & 72.35 & - & - & \textbf{54.26}\\
\hline
Ours (transfer) & This paper & Duke & \textbf{84.4} & \textbf{92.8} & \textbf{95.0} & \textbf{60.4} &Market & \underline{72.4} & \textbf{82.9}	& \underline{85.0} & 51.4 \\
\hline
\end{tabular}
}
\end{center}
\vspace{-2mm}
\caption{Unsupervised person re-ID performance comparison with state-of-the-art methods on Market-1501 and DukeMTMC-reID.}
\vspace{-2mm}
\label{table:sota-market-duke}
\end{table*}

\begin{table}
\footnotesize
\begin{center}
\begin{tabular}{l|c|c|c|c|c}
\hline
\multirow{2}{*}{Method} & \multirow{2}{*}{Source} & \multicolumn{4}{c}{MSMT17} \\
\cline{3-6}
& &Rank-1 & Rank-5 & Rank-10 & mAP  \\
\hline\hline
Ours & None & \textbf{35.4} & \textbf{44.8} & \textbf{49.8}& \textbf{11.2} \\
\hline\hline
PTGAN~\cite{wei2018person} & Market &10.2 & - & 24.4 & 2.9 \\
ECN~\cite{zhong2019invariance} & Market &  25.3 & 36.3 & 42.1 & 8.5 \\
SSG~\cite{Fu_2019_ICCV} & Market & 31.6 & - & 49.6 & 13.2\\
Ours (transfer) & Market & \textbf{40.8} & \textbf{51.8} & \textbf{56.7} & \textbf{15.1} \\
\hline
PTGAN~\cite{wei2018person} & Duke &11.8 & - & 27.4 & 3.3 \\
ECN~\cite{zhong2019invariance} & Duke & 30.2 & 41.5 & 46.8 & 10.2 \\
SSG~\cite{Fu_2019_ICCV} & Duke & 32.2 & - & 51.2 & 13.3\\
Ours (transfer) & Duke & \textbf{43.6} & \textbf{54.3} & \textbf{58.9} & \textbf{16.2}\\
\hline
\end{tabular}
\end{center}
\vspace{-2mm}
\caption{Comparison with state-of-the-art methods on MSMT17.}
\vspace{-2mm}
\label{table:sota-msmt17}
\end{table}

\vspace{-1mm}
\subsection{Comparison with the State of the Art}
We compare our method against state-of-the-art unsupervised learning and transfer learning approaches on Market-1501~\cite{zheng2015scalable}, DukeMTMC-reID~\cite{ristani2016performance} and MSMT17~\cite{wei2018person}. Table \ref{table:sota-market-duke} and Table \ref{table:sota-msmt17} summarize the comparison.

Table \ref{table:sota-market-duke} reports comparisons on Market-1501 and DukeMTMC-reID. We compare two types of methods, including unsupervised learning methods: LOMO~\cite{liao2015person}, BOW~\cite{zheng2015scalable},  BUC~\cite{lin2019bottom} and DBC~\cite{ding12dispersion}, and transfer learning based approaches: PUL~\cite{fan2018unsupervised}, PTGAN~\cite{wei2018person}, SPGAN~\cite{deng2018image}, CAMEL~\cite{yu2017cross}, MMFA~\cite{lin2018multi}, TJ-AIDL~\cite{wang2018transferable}, HHL~\cite{zhong2018generalizing}, ECN~\cite{zhong2019invariance}, MAR~\cite{yu2019unsupervised}, PAUL~\cite{Yang_2019_CVPR}, SSG~\cite{Fu_2019_ICCV}, CR-GAN~\cite{Chen_2019_ICCV}, CASCL~\cite{Wu_2019_ICCV}, PDA-Net~\cite{ Li_2019_ICCV}, UCDA~\cite{Qi_2019_ICCV} and PAST~\cite{Zhang_2019_ICCV}.

We first compare with unsupervised learning methods. LOMO and BOW utilize hand-crafted features, which show lower performance. BUC and DBC treat each image as a single cluster then merges clusters, thus share certain similarity to our work. However, our method outperforms them by large margins. The reasons could be because: 1) BUC tries to keep different clusters with similar size, hence could suffer from the issue of imbalanced number of positive labels. MPLP could alleviate this issue by adaptively selecting positive labels for different images. 2) As discussed in ablation studies, MMCL performs better than the cross entropy loss when using memory bank as classifier. Therefore, MPLP and MMCL effectively boost our performance.

Under the transfer learning setting, our method achieves the best performance on Market-1501 in Table \ref{table:sota-market-duke}. For example, our rank-1 accuracy of on Market-1501 achieves 84.4\%, when using DukeMTMC-reID as the source dataset. Similarly, we get 72.4\% rank-1 accuracy on DukeMTMC-reID using Market-1501 as the source dataset. Although SSG and PAST achieve slightly better performance, our method is more flexible and can be used without labeled dataset.  It is also interesting to observe that, with unsupervised learning setting, our method still outperforms several recent transfer learning methods, \emph{e.g.}, our rank-1 accuracy of 80.3\% on Market-1501 vs. 78.38\% and 64.3\% of PAST~\cite{Zhang_2019_ICCV} and UCDA~\cite{Qi_2019_ICCV}.

We also conduct experiments on MSMT17, a larger and more challenging dataset. A limited number of works report performance on MSMT17, \emph{i.e.}, PTGAN~\cite{wei2018person}, ECN~\cite{zhong2019invariance}, and SSG~\cite{Fu_2019_ICCV}. As table \ref{table:sota-msmt17} shows, our approach outperforms existing methods by large margins under both unsupervised and transfer learning settings. For example, our method achieves 35.4\% and 43.6\%/40.8\% rank-1 accuracy respectively. 
This outperforms SSG by 11.4\% in rank-1 accuracy. The above experiments on three datasets demonstrate the promising performance of our MPLP and MMCL.

\section{Conclusion}
This paper proposes a multi-label classification method to address unsupervised person ReID. Different from previous works, our method works without requiring any labeled data or a good pre-trained model. This is achieved by iteratively predicting multi-class labels and updating the network with a multi-label classification loss. MPLP is proposed for multi-class label prediction by considering both visual similarity and cycle consistency. MMCL is introduced to compute the multi-label classification loss and address the vanishing gradient issue. Experiments on several large-scale datasets demonstrate the effectiveness of the proposed methods in unsupervised person ReID.

\vspace{5mm}
\noindent\textbf{Acknowledgement} This work is supported in part by The National Key Research and Development Program of China under Grant No. 2018YFE0118400, in part by Beijing Natural Science Foundation under Grant No. JQ18012, in part by Natural Science Foundation of China under Grant No. 61936011, 61425025, 61620106009, 61572050, 91538111.

{\small
\bibliographystyle{ieee_fullname}
\bibliography{egbib}

\begin{thebibliography}{10}\itemsep=-1pt

\bibitem{Chen_2019_ICCV}
Yanbei Chen, Xiatian Zhu, and Shaogang Gong.
\newblock Instance-guided context rendering for cross-domain person
  re-identification.
\newblock In {\em ICCV}, 2019.

\bibitem{5206848}
J. {Deng}, W. {Dong}, R. {Socher}, L. {Li}, {Kai Li}, and {Li Fei-Fei}.
\newblock Imagenet: A large-scale hierarchical image database.
\newblock In {\em CVPR}, 2009.

\bibitem{deng2018image}
Weijian Deng, Liang Zheng, Qixiang Ye, Guoliang Kang, Yi Yang, and Jianbin
  Jiao.
\newblock Image-image domain adaptation with preserved self-similarity and
  domain-dissimilarity for person re-identification.
\newblock In {\em CVPR}, 2018.

\bibitem{ding12dispersion}
Guodong Ding, Salman Khan, Qingze Yin, and Zhenmin Tang.
\newblock Dispersion based clustering for unsupervised person
  re-identification.
\newblock In {\em BMVC}, 2019.

\bibitem{Durand_2019_CVPR}
Thibaut Durand, Nazanin Mehrasa, and Greg Mori.
\newblock Learning a deep convnet for multi-label classification with partial
  labels.
\newblock In {\em CVPR}, 2019.

\bibitem{fan2018unsupervised}
Hehe Fan, Liang Zheng, Chenggang Yan, and Yi Yang.
\newblock Unsupervised person re-identification: Clustering and fine-tuning.
\newblock {\em ACM Transactions on Multimedia Computing, Communications, and
  Applications (TOMM)}, 14(4):83, 2018.

\bibitem{Fu_2019_ICCV}
Yang Fu, Yunchao Wei, Guanshuo Wang, Yuqian Zhou, Honghui Shi, and Thomas~S.
  Huang.
\newblock Self-similarity grouping: A simple unsupervised cross domain
  adaptation approach for person re-identification.
\newblock In {\em ICCV}, 2019.

\bibitem{gutmann2010noise}
Michael Gutmann and Aapo Hyv{\"a}rinen.
\newblock Noise-contrastive estimation: A new estimation principle for
  unnormalized statistical models.
\newblock In {\em AISTATS}, 2010.

\bibitem{he2016deep}
Kaiming He, Xiangyu Zhang, Shaoqing Ren, and Jian Sun.
\newblock Deep residual learning for image recognition.
\newblock In {\em CVPR}, 2016.

\bibitem{ioffe2015batch}
Sergey Ioffe and Christian Szegedy.
\newblock Batch normalization: Accelerating deep network training by reducing
  internal covariate shift.
\newblock In {\em ICML}, 2015.

\bibitem{iscen2018mining}
Ahmet Iscen, Giorgos Tolias, Yannis Avrithis, and Ond{\v{r}}ej Chum.
\newblock Mining on manifolds: Metric learning without labels.
\newblock In {\em CVPR}, 2018.

\bibitem{jegou2007contextual}
Herve Jegou, Hedi Harzallah, and Cordelia Schmid.
\newblock A contextual dissimilarity measure for accurate and efficient image
  search.
\newblock In {\em CVPR}, 2007.

\bibitem{komodakis2018unsupervised}
Nikos Komodakis and Spyros Gidaris.
\newblock Unsupervised representation learning by predicting image rotations.
\newblock In {\em ICLR}, 2018.

\bibitem{krizhevsky2012imagenet}
Alex Krizhevsky, Ilya Sutskever, and Geoffrey~E Hinton.
\newblock Imagenet classification with deep convolutional neural networks.
\newblock In {\em NeurIPS}, 2012.

\bibitem{li2018unsupervised}
Junnan Li, Yongkang Wong, Qi Zhao, and Mohan Kankanhalli.
\newblock Unsupervised learning of view-invariant action representations.
\newblock In {\em NeurIPS}, 2018.

\bibitem{li2019pose}
Jianing Li, Shiliang Zhang, Qi Tian, Meng Wang, and Wen Gao.
\newblock Pose-guided representation learning for person re-identification.
\newblock {\em IEEE Trans. Pattern Anal. Mach. Intell.}, 2019.

\bibitem{Li_2019_ICCV}
Yu-Jhe Li, Ci-Siang Lin, Yan-Bo Lin, and Yu-Chiang~Frank Wang.
\newblock Cross-dataset person re-identification via unsupervised pose
  disentanglement and adaptation.
\newblock In {\em ICCV}, 2019.

\bibitem{liao2015person}
Shengcai Liao, Yang Hu, Xiangyu Zhu, and Stan~Z Li.
\newblock Person re-identification by local maximal occurrence representation
  and metric learning.
\newblock In {\em CVPR}, 2015.

\bibitem{lin2018multi}
Shan Lin, Haoliang Li, Chang-Tsun Li, and Alex~Chichung Kot.
\newblock Multi-task mid-level feature alignment network for unsupervised
  cross-dataset person re-identification.
\newblock In {\em BMVC}, 2018.

\bibitem{lin2019bottom}
Yutian Lin, Xuanyi Dong, Liang Zheng, Yan Yan, and Yi Yang.
\newblock A bottom-up clustering approach to unsupervised person
  re-identification.
\newblock In {\em AAAI}, 2019.

\bibitem{long2015learning}
Mingsheng Long, Yue Cao, Jianmin Wang, and Michael~I Jordan.
\newblock Learning transferable features with deep adaptation networks.
\newblock {\em arXiv preprint arXiv:1502.02791}, 2015.

\bibitem{Lv_2018_CVPR}
Jianming Lv, Weihang Chen, Qing Li, and Can Yang.
\newblock Unsupervised cross-dataset person re-identification by transfer
  learning of spatial-temporal patterns.
\newblock In {\em CVPR}, 2018.

\bibitem{morin2005hierarchical}
Frederic Morin and Yoshua Bengio.
\newblock Hierarchical probabilistic neural network language model.
\newblock In {\em Aistats}, 2005.

\bibitem{Qi_2019_ICCV}
Lei Qi, Lei Wang, Jing Huo, Luping Zhou, Yinghuan Shi, and Yang Gao.
\newblock A novel unsupervised camera-aware domain adaptation framework for
  person re-identification.
\newblock In {\em ICCV}, 2019.

\bibitem{ristani2016performance}
Ergys Ristani, Francesco Solera, Roger Zou, Rita Cucchiara, and Carlo Tomasi.
\newblock Performance measures and a data set for multi-target, multi-camera
  tracking.
\newblock In {\em ECCV}, 2016.

\bibitem{su2017pose}
Chi Su, Jianing Li, Shiliang Zhang, Junliang Xing, Wen Gao, and Qi Tian.
\newblock Pose-driven deep convolutional model for person re-identification.
\newblock In {\em ICCV}, 2017.

\bibitem{su2015multi}
Chi Su, Fan Yang, Shiliang Zhang, Qi Tian, Larry~S Davis, and Wen Gao.
\newblock Multi-task learning with low rank attribute embedding for person
  re-identification.
\newblock In {\em ICCV}, 2015.

\bibitem{su2016deep}
Chi Su, Shiliang Zhang, Junliang Xing, Wen Gao, and Qi Tian.
\newblock Deep attributes driven multi-camera person re-identification.
\newblock In {\em ECCV}, 2016.

\bibitem{su2017attributes}
Chi Su, Shiliang Zhang, Fan Yang, Guangxiao Zhang, Qi Tian, Wen Gao, and
  Larry~S Davis.
\newblock Attributes driven tracklet-to-tracklet person re-identification using
  latent prototypes space mapping.
\newblock {\em Pattern Recognition}, 66:4--15, 2017.

\bibitem{wang2018transferable}
Jingya Wang, Xiatian Zhu, Shaogang Gong, and Wei Li.
\newblock Transferable joint attribute-identity deep learning for unsupervised
  person re-identification.
\newblock In {\em CVPR}, 2018.

\bibitem{wei2018person}
Longhui Wei, Shiliang Zhang, Wen Gao, and Qi Tian.
\newblock Person transfer gan to bridge domain gap for person
  re-identification.
\newblock In {\em CVPR}, 2018.

\bibitem{Wu_2019_ICCV}
Ancong Wu, Wei-Shi Zheng, and Jian-Huang Lai.
\newblock Unsupervised person re-identification by camera-aware similarity
  consistency learning.
\newblock In {\em ICCV}, 2019.

\bibitem{wu2017sampling}
Chao-Yuan Wu, R Manmatha, Alexander~J Smola, and Philipp Krahenbuhl.
\newblock Sampling matters in deep embedding learning.
\newblock In {\em CVPR}, 2017.

\bibitem{wu2018unsupervised}
Zhirong Wu, Yuanjun Xiong, Stella~X Yu, and Dahua Lin.
\newblock Unsupervised feature learning via non-parametric instance
  discrimination.
\newblock In {\em CVPR}, 2018.

\bibitem{Yan_2017_CVPR}
Hongliang Yan, Yukang Ding, Peihua Li, Qilong Wang, Yong Xu, and Wangmeng Zuo.
\newblock Mind the class weight bias: Weighted maximum mean discrepancy for
  unsupervised domain adaptation.
\newblock In {\em CVPR}, 2017.

\bibitem{Yang_2019_CVPR}
Qize Yang, Hong-Xing Yu, Ancong Wu, and Wei-Shi Zheng.
\newblock Patch-based discriminative feature learning for unsupervised person
  re-identification.
\newblock In {\em CVPR}, 2019.

\bibitem{yu2017cross}
Hong-Xing Yu, Ancong Wu, and Wei-Shi Zheng.
\newblock Cross-view asymmetric metric learning for unsupervised person
  re-identification.
\newblock In {\em CVPR}, 2017.

\bibitem{yu2019unsupervised}
Hong-Xing Yu, Wei-Shi Zheng, Ancong Wu, Xiaowei Guo, Shaogang Gong, and
  Jian-Huang Lai.
\newblock Unsupervised person re-identification by soft multilabel learning.
\newblock In {\em CVPR}, 2019.

\bibitem{zhang2013review}
Min-Ling Zhang and Zhi-Hua Zhou.
\newblock A review on multi-label learning algorithms.
\newblock {\em IEEE transactions on knowledge and data engineering},
  26(8):1819--1837, 2013.

\bibitem{Zhang_2019_ICCV}
Xinyu Zhang, Jiewei Cao, Chunhua Shen, and Mingyu You.
\newblock Self-training with progressive augmentation for unsupervised
  cross-domain person re-identification.
\newblock In {\em ICCV}, 2019.

\bibitem{NIPS2018_8094}
Zhilu Zhang and Mert Sabuncu.
\newblock Generalized cross entropy loss for training deep neural networks with
  noisy labels.
\newblock In {\em NeurIPS}. 2018.

\bibitem{zheng2015scalable}
Liang Zheng, Liyue Shen, Lu Tian, Shengjin Wang, Jingdong Wang, and Qi Tian.
\newblock Scalable person re-identification: A benchmark.
\newblock In {\em ICCV}, 2015.

\bibitem{zheng2016person}
Liang Zheng, Yi Yang, and Alexander~G Hauptmann.
\newblock Person re-identification: Past, present and future.
\newblock {\em arXiv preprint arXiv:1610.02984}, 2016.

\bibitem{zhong2017re}
Zhun Zhong, Liang Zheng, Donglin Cao, and Shaozi Li.
\newblock Re-ranking person re-identification with k-reciprocal encoding.
\newblock In {\em CVPR}, 2017.

\bibitem{zhong2018generalizing}
Zhun Zhong, Liang Zheng, Shaozi Li, and Yi Yang.
\newblock Generalizing a person retrieval model hetero-and homogeneously.
\newblock In {\em ECCV}, 2018.

\bibitem{zhong2019invariance}
Zhun Zhong, Liang Zheng, Zhiming Luo, Shaozi Li, and Yi Yang.
\newblock Invariance matters: Exemplar memory for domain adaptive person
  re-identification.
\newblock In {\em CVPR}, 2019.

\bibitem{zhong2018camera}
Zhun Zhong, Liang Zheng, Zhedong Zheng, Shaozi Li, and Yi Yang.
\newblock Camera style adaptation for person re-identification.
\newblock In {\em CVPR}, 2018.

\end{thebibliography}
}

\end{document}